\documentclass[runningheads]{llncs}
\usepackage{graphicx}
\usepackage{comment}
\usepackage{amsmath,amssymb} % define this before the line numbering.
\usepackage{color}
\usepackage{multirow}
\usepackage{booktabs}
\usepackage{wrapfig}
\usepackage{hyperref}
\usepackage[misc]{ifsym}

\usepackage{array}
\newcommand{\PreserveBackslash}[1]{\let\temp=\\#1\let\\=\temp}
\newcolumntype{C}[1]{>{\PreserveBackslash\centering}p{#1}}
\newcolumntype{R}[1]{>{\PreserveBackslash\raggedleft}p{#1}}
\newcolumntype{L}[1]{>{\PreserveBackslash\raggedright}p{#1}}

\begin{document}

% \renewcommand\thelinenumber{\color[rgb]{0.2,0.5,0.8}\normalfont\sffamily\scriptsize\arabic{linenumber}\color[rgb]{0,0,0}}
% \renewcommand\makeLineNumber {\hss\thelinenumber\ \hspace{6mm} \rlap{\hskip\textwidth\ \hspace{6.5mm}\thelinenumber}}
% \linenumbers
\pagestyle{headings}
\mainmatter
\def\ECCVSubNumber{622}  % Insert your submission number here

% \title{Towards Incorporating Temporal Prior for Action Localization} % Replace with your title
\title{Bottom-Up Temporal Action Localization with Mutual Regularization}

% INITIAL SUBMISSION 
%\begin{comment}
% \titlerunning{ECCV-20 submission ID \ECCVSubNumber} 
% \authorrunning{ECCV-20 submission ID \ECCVSubNumber} 
% \author{Anonymous ECCV submission}
% \institute{Paper ID \ECCVSubNumber}
%\end{comment}
%******************

% CAMERA READY SUBMISSION
% \begin{comment}
\titlerunning{Bottom-Up Temporal Action Localization with Mutual Regularization}
% If the paper title is too long for the running head, you can set
% an abbreviated paper title here
%
\author{Peisen Zhao\inst{1} \and %\orcidID{0000-0002-3787-193X} \and
Lingxi Xie\inst{2} \and %\orcidID{1111-2222-3333-4444} \and
Chen Ju\inst{1} \and %\orcidID{2222--3333-4444-5555}
Ya Zhang\inst{1}\textsuperscript{\Letter} \and %\orcidID{2222--3333-4444-5555}
Yanfeng Wang\inst{1} \and \\ %\orcidID{2222--3333-4444-5555}
Qi Tian\inst{2}}%\orcidID{2222--3333-4444-5555}}
\authorrunning{P. Zhao et al.}
% First names are abbreviated in the running head.
% If there are more than two authors, 'et al.' is used.
%
\institute{Cooperative Medianet Innovation Center, Shanghai Jiao Tong University\\
\email{\{pszhao, ju\_chen, ya\_zhang, wangyanfeng\}@sjtu.edu.cn}\\
\and
Huawei Inc.\\
\email{198808xc@gmail.com, tian.qi1@huawei.com}}
% \end{comment}
%******************
\maketitle

\begin{abstract}

Recently, temporal action localization (TAL), \textit{i.e.}, finding specific action segments in untrimmed videos, has attracted increasing attentions of the computer vision community. State-of-the-art solutions for TAL involves evaluating the frame-level probabilities of three action-indicating phases, \textit{i.e.} starting, continuing, and ending; and then post-processing these predictions for the final localization. This paper delves deep into this mechanism, and argues that existing methods, by modeling these phases as individual classification tasks, ignored the potential temporal constraints between them. This can lead to incorrect and/or inconsistent predictions when some frames of the video input lack sufficient discriminative information. To alleviate this problem, we introduce two regularization terms to mutually regularize the learning procedure: the Intra-phase Consistency (IntraC) regularization is proposed to make the predictions verified inside each phase; and the Inter-phase Consistency (InterC) regularization is proposed to keep consistency between these phases. Jointly optimizing these two terms, the entire framework is aware of these potential constraints during an end-to-end optimization process. Experiments are performed on two popular TAL datasets, THUMOS14 and ActivityNet1.3. Our approach clearly outperforms the baseline both quantitatively and qualitatively. The proposed regularization also generalizes to other TAL methods (\textit{e.g.}, TSA-Net and PGCN). code: \url{https://github.com/PeisenZhao/Bottom-Up-TAL-with-MR}

\end{abstract}

%%%%%%%%% BODY TEXT
\section{Introduction}

Temporal Action Localization (TAL), aiming to locate action instances from untrimmed videos, is a fundamental task in video content analysis. 
TAL can be divided into two parts, temporal action proposal and action classification. The latter is relatively well studied with cogent performance achieved by recent action classifiers  \cite{carreira2017quo,qiu2017learning,tran2018closer,xie2018rethinking,wang2018appearance}. To improve the performance in standard benchmarks \cite{jiang2014thumos,caba2015activitynet}, how to generate precise action proposals remains a challenge.

Early approaches for generating action proposals mostly adopt a \textbf{top-down} approach, i.e., first generate regularly distributed proposals (\emph{e.g.}, multi-scale sliding windows), and then evaluate their confidence. However, the top-down methods \cite{caba2016fast,escorcia2016daps,shou2016temporal,buch2017sst,gao2017turn} often suffer from over-generating candidate proposals and rigid proposal boundaries.
To solve the above problem, 
\textbf{bottom-up} approaches have been proposed~\cite{xiong2017pursuit,lin2018bsn,lin2019bmn,liu2019multi,gong2019scale}.
A typical bottom-up method first densely evaluates the frame-level probabilities of three action-indicating phases, i.e. starting, continuing, and ending; then groups action proposals based on the located candidate starting and ending points. This design paradigm enables flexible action proposal generation and achieves a high recall with fewer proposals \cite{xiong2017pursuit}, which has become a more preferred practice in temporal action proposals.

Predicting the frame-level probability of the starting, continuing, and ending phases of actions is crucial for the success of bottom-up approaches.
Existing methods model it as three binary classification tasks and use frame-level positive and negative labels converted from action temporal location as supervision, which can suffer the difficulty of learning from limited and/or ambiguous training data. 
In particular, it is often difficult to determine the accurate time that an action starts or ends, and even when the action continues, there is no guarantee that every frame contains sufficient information of being correctly classified. 
In other words, one may need to refer to complementary information to judge the status of an action, \textit{e.g.}, if there is no clear sign that an action has ended, the probability that it is continuing is high. 
Ignoring such temporal relationship may lead to erroneous and inconsistent predictions.
Thus, independent classification tasks have the following two drawbacks.
\textbf{First}, each temporal location is considered as an isolated instance and their probabilities are calculated independently, without considering the temporal relationship among them. In fact, for any of the three phases, the probability is expected to have relatively smooth predictions among contiguous temporal locations. Ignoring the temporal relationship may leads to inconsistent predictions.
\textbf{Second}, the modeling of the probability for starting, continuing, and ending phases are independent of each other. In fact, for any action, the starting, continuing, and ending phases always come as an ordered triplet. Ignoring the ordering relationship of the three phases could lead to contradictory predictions.

In this paper, we address this issue explicitly by exploring two regularization terms.
To enforces the temporal relationship among predictions,  \textbf{Intra-phase Consistency} (IntraC) regularization is proposed, which targets to minimize the discrepancy inside \textit{positive} or \textit{negative} regions of each phases, and maximize the discrepancy between \textit{positive} and  \textit{negative} regions.
To meet the ordering constraint of the three phases, we introduce \textbf{Inter-phase Consistency} (InterC) regularization, which enforces consistency among the probability of the three phases, by operating between continuing-starting and continuing-ending.
When introducing the above two regularization terms to the original loss of bottom-up temporal action localization network, the optimization of IntraC and InterC 
may be considered as a form of 
mutual regularization among the three classifiers, since the  predictions of the three phases are now coupled via consistency check on classifier outputs.
With the above mutual regularization, the entire framework remains end-to-end trainable while enforcing the above constraints.

To validate the effectiveness of the proposed method, we perform experiments on two popular benchmark datasets, THUMOS14 and ActivityNet1.3. Our experimental results have demonstrated that our approach clearly outperforms the state-of-the-arts both quantitatively and qualitatively. Especially on THUMOS14 dataset, we improve absolute $6.8\%$ mAP at a strict IoU of $0.7$ settings from the previous best. Moreover, we show that the proposed mutual regularization is independent of the temporal action localization framework. When we introduce IntraC and InterC to other network (TSA-Net \cite{gong2019scale}) or framework (PGCN \cite{zeng2019graph}), better performance is also achieved.

\section{Related Work}

\noindent \textbf{Action recognition}. 
Same as image recognition in image analysis, action recognition is a fundamental task in video domain. Extensive models \cite{tran2015learning,carreira2017quo,tran2018closer,xie2018rethinking,wang2018appearance,zhao2020u2s} on action recognition have been widely studied. Deeper models \cite{carreira2017quo,qiu2017learning,lin2019tsm}, more massive datasets \cite{karpathy2014large,abu2016youtube,kay2017kinetics,monfort2019moments}, and smarter supervision~\cite{gan2016webly,gan2016you} have promoted the development of this direction. These action recognition approaches are based on trimmed videos, which are not suitable for untrimmed videos due to the considerable duration of the background. However, the pre-trained models on action recognition task can provide effective feature representation for temporal action localization task. In this paper, we use the I3D model \cite{carreira2017quo}, pre-trained on Kinetics \cite{kay2017kinetics}, to extract video features.

\noindent \textbf{Temporal action localization}. 
Temporal action localization is a mirror problem of image object detection\cite{ren2015faster,redmon2016you} in the temporal domain. The TAL task can be decomposed into proposal generation and classification stage, same as the two-stage approach of object detection. Recent methods for proposal generation are divided into two branches, top-down and bottom-up fashions. Top-down approaches \cite{buch2017sst,caba2016fast,escorcia2016daps,gao2017turn,shou2016temporal,dai2017temporal,xu2017rc3d,chao2018rethinking} generated proposals with pre-defined regularly distributed segments then evaluated the confidence of each proposal. The boundary of top-down proposals are not flexible, and these generation strategies often cause extensive false positive proposals, which will introduce burdens in the classification stage. However, the other bottom-up approaches alleviated this problem and achieved the new state-of-the-art. TAG~\cite{xiong2017pursuit} was an early study of bottom-up fashion, which used frame-level action probabilities to group action proposals. Lin \emph{et al.} proposed the multi-stage BSN \cite{lin2018bsn} and end-to-end BMN \cite{lin2019bmn} models via locating temporal boundaries to generate action proposals. \emph{Gong al.} \cite{gong2019scale} also predicted action probabilities to generate action proposals from the perspective of multi scales. Zeng \emph{et al.} proposed the PGCN \cite{zeng2019graph} to model the proposal-proposal relations based on bottom-up proposals. Combined top-down and bottom-up fashions, Liu \emph{et al.} proposed a MGG \cite{liu2019multi} model, which takes advantage of frame-level action probability as well. \cite{yuan2017temporal} is relevant to out study that enforced the temporal structure by maximizing the top-K summation of the confidence scores of the starting, continuing, and ending.

\section{Method}

\begin{figure*}[t]
\begin{center}
  \includegraphics[width=1.0\linewidth]{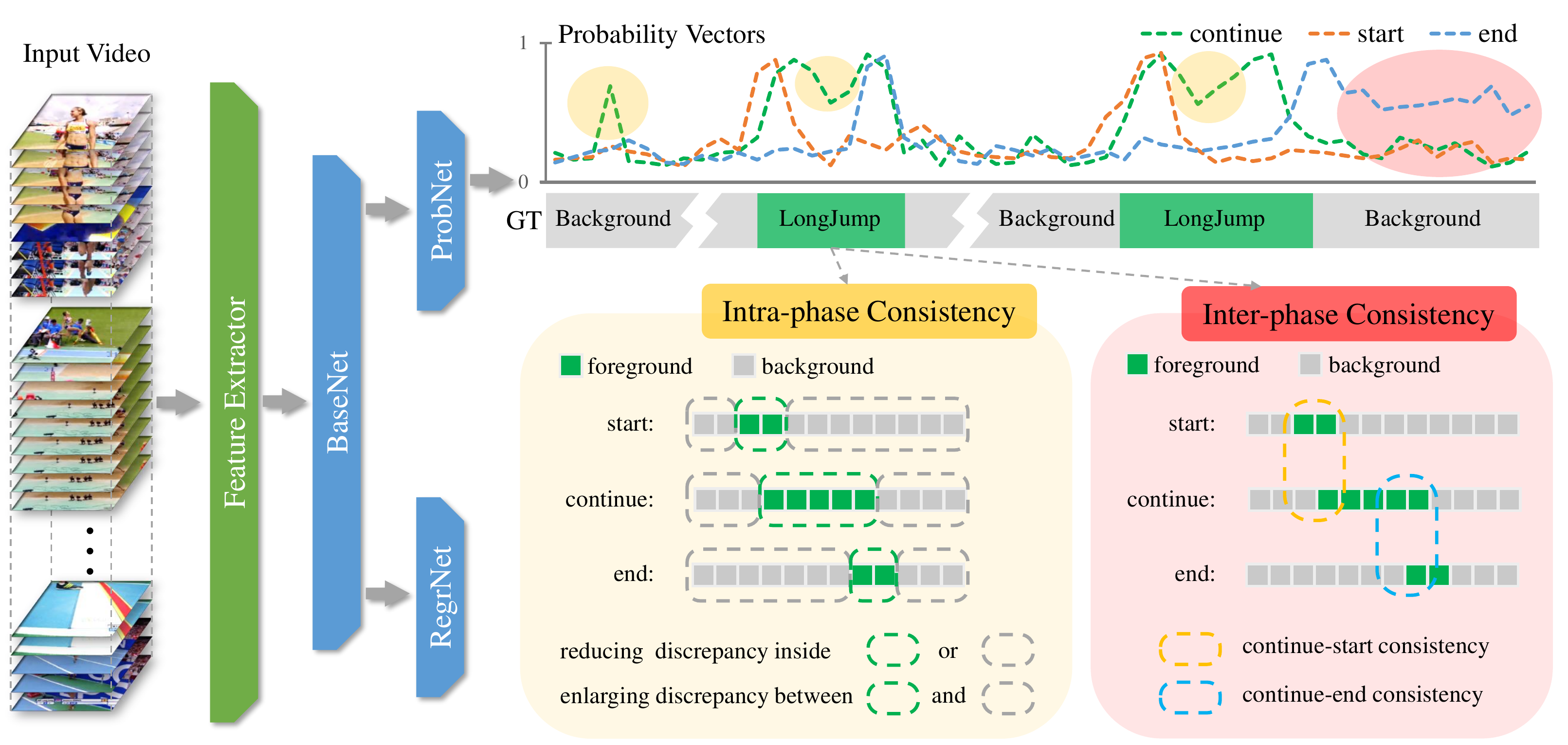}
\end{center}
\caption{Schematic of our approach. Three probability phases are predicted by the ProbNet. Intra-phase Consistency loss is built inside each phase by first separating \textit{positive} and \textit{negative} regions, then reduce the discrepancy inside \textit{positive} or \textit{negative}, and enlarge the discrepancy between \textit{positive} and \textit{negative}. Inter-phase Consistency loss is built between the continue-start phase and the continue-end phase.}
\label{fig:framework}
\end{figure*}

\subsection{Problem and Baseline}

\noindent \textbf{Notations.} Given an Untrimmed video, we denote $\{\mathbf{f}_{t}\}_{t=1}^{T}$ as a feature sequence to represent a video, where $T$ is the length of the video and $\mathbf{f}_{t}$ is the $t$-th feature vector extracted from continuous RGB frames and optical flows. Annotations are $\mathcal{\varphi}=\{(t_{s,n},t_{e,n},a_{n})\}_{n=1}^{N}$, where $t_{s,n}$, $t_{e,n}$, and $a_{n}$ are start time, end time, and class label of the action instance $n$. $N$ is the number of action annotations. Following previous studies \cite{lin2018bsn,lin2019bmn,liu2019multi,gong2019scale}, we predict continuing, starting, and ending probability vectors $\mathbf{p}^\mathrm{C}\in [0,1]^{T}$, $\mathbf{p}^\mathrm{S}\in [0,1]^{T}$, and $\mathbf{p}^\mathrm{E}\in [0,1]^{T}$ to generate action proposals. Correspondingly, the ground-truth labels are generated via $\mathcal{\varphi}$, which are notated by $\mathbf{g}^\mathrm{C}\in \{0,1\}^{T}$, $\mathbf{g}^\mathrm{S}\in \{0,1\}^{T}$, and $\mathbf{g}^\mathrm{E}\in \{0,1\}^{T}$, respectively. Continuing ground-truth $\mathbf{g}^\mathrm{C}$ has value ``1'' inside the action instances $[t_{s,n},t_{e,n}]$, while starting and ending points are expanded to a region $[t_{s,n}-\delta_{n},t_{s,n}+\delta_{n}]$ and $[t_{e,n}-\delta_{n},t_{e,n}+\delta_{n}]$ to assign the ground-truth label $\mathbf{g}^\mathrm{S}$ and $\mathbf{g}^\mathrm{E}$. $\delta_{n}$ is set to be 0.1 duration of the action instance $n$, same as \cite{lin2018bsn,lin2019bmn,gong2019scale}.

\noindent \textbf{Baseline.} This paper takes the typical bottom-up TAL framework as our baseline, such as BSN~\cite{lin2018bsn}. As illustrated in Figure~\ref{fig:framework}, the baseline network is trained without Intra-phase Consistency and Inter-phase Consistency. We first use 3D convolutional network to extract video features $\{\mathbf{f}_{t}\}_{t=1}^{T}$, then feed the feature sequence to several 1D convolutional networks to (i) predict three probability vectors ($\mathbf{p}^\mathrm{C}$, $\mathbf{p}^\mathrm{S}$, and $\mathbf{p}^\mathrm{E}$) by ProbNet, (ii) predict the starting and ending boundary offsets ($\hat{\mathbf{o}}_\mathrm{S}$ and $\hat{\mathbf{o}}_\mathrm{E}$) by RegrNet. Finally, we generate proposals by combining start-end pairs with high probabilities and classify these candidate proposals.

\subsection{Motivation: Avoiding Ambiguity with Temporal Consistency}

The first and fundamental procedure in bottom-up TAL is to predict frame-level probabilities of three action-indicating phases, \textit{i.e.} starting, continuing, and ending. 
Existing approaches use frame-level labels, $\mathbf{g}^\mathrm{C}\in \{0,1\}^{T}$, $\mathbf{g}^\mathrm{S}\in \{0,1\}^{T}$, and $\mathbf{g}^\mathrm{E}\in \{0,1\}^{T}$ to train three binary classification tasks. 
Since the meaning of ``starting'', ``continuing'', and ``ending'' have certain ambiguity, it is hard to determine the accurate time that an action starts, ends, and continues. Moreover, we find that even in training set, the \textbf{False Alarm} of these binary classification tasks reaches 68\%, 64\%, and 28\% for starting, ending, and continuing, respectively. 
As shown in Figure~\ref{fig:framework}, we can also observe that the continuing phase in green are not stable inside an action instance ``LongJump'' or background (\textit{yellow} circles); and different action phases are not support each other (\textit{red} circle).
Thus, only supervised by classification labels is hard to optimize these problem, because there is no guarantee that every frame contains sufficient information of being correctly classified.

Therefore, to better regularize the learning process of avoiding ambiguity, we propose two consistency regularization terms during an end-to-end optimization, that consider the relations between different temporal locations inside each probability phase, named \textbf{Intra-phase Consistency} (IntraC) and the relations among different probability phases, named \textbf{Inter-phase Consistency} (InterC).

\subsection{Adding Mutual Regularization}

As illustrated in Figure~\ref{fig:framework}, we add two consistency losses, IntraC and InterC, to regularize the learning process. IntraC is built inside each phase by first separating \textit{positive} and \textit{negative} regions, then reduce the discrepancy inside \textit{positive} or \textit{negative}, and enlarge the discrepancy between \textit{positive} and \textit{negative}. InterC performs consistency among three phases, which operates between continuing-starting and continuing-ending, \textbf{(i)} if there were an abrupt rise in the continuing phase, the starting phase should give a high probability, and vise versa; \textbf{(ii)} if there were an abrupt drop in the continuing phase, the ending phase should give a high probability, and vise versa.

\noindent\textbf{Intra-phase Consistency.} We build our Intra-phase Consistency loss inside each per-frame probability phase of start, end, and continuing. Firstly, we show the detailed operations for continuing phase $\mathbf{p}^\mathrm{C}$. The yellow block in Figure~\ref{fig:framework} shows an example of the IntraC on continuing phase $\mathbf{p}^\mathrm{C}$. To make the per-frame predictions supervised by their context predictions, we first define the \textit{positive} and \textit{negative} regions. The \textit{positive} regions are defined as the locations where action continues by $g_t^\mathrm{C}=1$, and the \textit{negative} regions are the rest of the time where $g_t^\mathrm{C}=0$. 
In terms of the division of the \textit{positive} and \textit{negative} region, the predicted continuing probabilities $\{p_t^\mathrm{C}\}_{t=1}^{T}$ are divided into a positive set $\mathcal{U}^\mathrm{C}=\{p_{t}^\mathrm{C}\mid g_t^\mathrm{C}=1\}$ and a negative set $\mathcal{V}^\mathrm{C}=\{p_{t}^\mathrm{C}\mid g_t^\mathrm{C}=0\}$. 
To make each prediction is not only supervised by its own label but other context labels, we optimize this problem by \textbf{(i)} $\min f(p_{i}^\mathrm{C},p_{j}^\mathrm{C}), \forall p_{i}^\mathrm{C}\in\mathcal{U}^\mathrm{C}. \forall p_{j}^\mathrm{C}\in\mathcal{U}^\mathrm{C}$ \textbf{(ii)} $\max f(p_{i}^\mathrm{C},p_{j}^\mathrm{C}), \forall p_{i}^\mathrm{C}\in\mathcal{U}^\mathrm{C}. \forall p_{j}^\mathrm{C}\in\mathcal{V}^\mathrm{C}$,
where $f$ is a distance function ($l_1$ distance in our experiments) to measure the difference between $p_{i}^{\mathrm{C}}$ and $p_{j}^{\mathrm{C}}$.
Therefore, the IntraC on continuing probability phase $\mathbf{p}^\mathrm{C}$ is formulated in Eq.~\eqref{eq:IC2}:
\begin{equation}
\label{eq:IC2}
  \mathcal{L}_{\mathrm{Intra}^\mathrm{C}} = \frac{1}{N_{\mathrm{U}}}\sum_{i,j}(\mathbf{A}\odot \mathbf{M}_{\mathrm{U}})_{i,j} + \frac{1}{N_{\mathrm{V}}}\sum_{i,j}(\mathbf{A}\odot \mathbf{M}_{\mathrm{V}})_{i,j}
  + (1 - \frac{1}{N_{\mathrm{UV}}}\sum_{i,j}(\mathbf{A}\odot \mathbf{M}_{\mathrm{UV}})_{i,j}),
\end{equation}
where $\mathbf{A} \in [0,1]^{T\times T}$ is an adjacency matrix  to establish the relationship between predicted probabilities by measuring the distance between them. The elements in $\mathbf{A}$ are formulated as $a_{i,j} = f(p_{i}^{\mathrm{C}},p_{j}^{\mathrm{C}})$. $\mathbf{M}_{\mathrm{U}}$, $\mathbf{M}_{\mathrm{V}}$, and $\mathbf{M}_{\mathrm{UV}} \in \{0,1\}^{T\times T}$ are three masks to select the corresponding pairs $a_{i,j}$ in adjacency matrix $\mathbf{A}$ from $\mathcal{U}^\mathrm{C}$ set, $\mathcal{V}^\mathrm{C}$ set, and between $\mathcal{U}^\mathrm{C}$ and $\mathcal{V}^\mathrm{C}$ sets, respectively. The constants $N_{\mathrm{U}}$, $N_{\mathrm{V}}$, and $N_{\mathrm{UV}}$ represent the number of ``1'' in each mask matrix. $\odot$ stand for the element-wise product. 

Following this intra consistency between different frame-predictions, we reduce the discrepancy inside \textit{positive} or \textit{negative}, and enlarge the discrepancy between them. Replicating IntraC loss on continuing phase, we can also obtain the $\mathcal{L}_{\mathrm{IC}^\mathrm{S}}$ and $\mathcal{L}_{\mathrm{IC}^\mathrm{E}}$. Hence, the whole IntraC loss is formulated in Eq.~\eqref{eq:IC3}:
\begin{equation}\label{eq:IC3}
  \mathcal{L}_{\mathrm{Intra}} = \mathcal{L}_{\mathrm{Intra}^\mathrm{C}} + \mathcal{L}_{\mathrm{Intra}^\mathrm{S}} + \mathcal{L}_{\mathrm{Intra}^\mathrm{E}}.
\end{equation}

\noindent\textbf{Inter-phase Consistency.} We build our Inter-phase Consistency loss between three probability phases, continuing phase $\mathbf{p}^\mathrm{C}$, starting phase $\mathbf{p}^\mathrm{S}$, and ending phase $\mathbf{p}^\mathrm{E}$. To make the consistency between these probability phases, we propose two hypotheses, (i) if there were an abrupt rise in the continuing phase, the starting phase should give a high probability, and vise versa; (ii) if there were an abrupt drop in the continuing phase, the ending phase should give a high probability, and vise versa. Following these hypotheses, we use the first difference term of $\mathbf{p}^\mathrm{C}$ to capture the abrupt rise and drop of the continuing probability phase: $\mathrm{\Delta} \mathbf{p}^\mathrm{C} = p_{t+1}^\mathrm{C}-p_t^\mathrm{C}.$

As illustrated in red block of Figure~\ref{fig:framework}, we build two kinds of constraints for InterC, the continue-start constraint in yellow circle and the continue-end constraint in blue circle. We use the positive values in $\mathrm{\Delta} \mathbf{p}^\mathrm{C}$ to represent continuing probability rise rate, notated as $p_{t}^{+}=\max\{0,\mathrm{\Delta}{p}_{t}^\mathrm{C}\}$, and use negative values in $\mathrm{\Delta} \mathbf{p}^\mathrm{C}$ to represent continuing probability drop rate, notated as $p_{t}^{-}=-\min\{0,\mathrm{\Delta}{p}_{t}^\mathrm{C}\}$. Thus, to make predictions of continuing, starting, and ending support each other, we optimize this problem by \textbf{(i)} $\min f(p_{t}^{+},p_{t}^{\mathrm{S}})$ and \textbf{(ii)} $\min f(p_{t}^{-},p_{t}^{\mathrm{E}})$, where $f$ is a distance function ($l_1$ distance in our experiments) to measure the distance. Then the InterC is formulated in Eq.~\eqref{eq:EC3}:

\begin{equation}\label{eq:EC3}
  \mathcal{L}_{\mathrm{Inter}} = \frac{1}{T}\sum_{t=1}^{T} \mid p_{t}^{+} - p_{t}^\mathrm{S}\mid + \mid p_{t}^{-} - p_{t}^\mathrm{E} \mid.
\end{equation}

\noindent\textbf{Loss function.} Predicting continuing, starting, and ending probabilities are trained with the cross-entropy loss. We separate the calculation by the \textit{positive} and \textit{negative} regions; then mix them with a ratio of $1$:$1$ to balance the proportion of the \textit{positive} and the \textit{negative}. The loss of predicting the continuing probability is formulated in Eq.~\eqref{eq:loss1}:

\begin{equation}\label{eq:loss1}
  \mathcal{L}_{\mathrm{C}}=\frac{1}{T_{\mathrm{C}}^{+}}\sum_{t\in\mathcal{U}^\mathrm{C}}\ln(p_{t}^\mathrm{C})+\frac{1}{T_{\mathrm{C}}^{-}}\sum_{t\in\mathcal{V}^\mathrm{C}}\ln(1-p_{t}^\mathrm{C}),
\end{equation}
where $\mathcal{U}^\mathrm{C}$ and $\mathcal{V}^\mathrm{C}$ denote the \textit{positive} and \textit{negative} set in $\mathbf{p}^\mathrm{C}$, while $T_\mathrm{C}^{+}$ and $T_\mathrm{C}^{-}$ are the number of them, respectively. Replacing the script ``$\mathrm{C}$'' with ``$\mathrm{S}$'' or ``$\mathrm{E}$'' in Eq.~\eqref{eq:loss1}, we can obtain the $\mathcal{L}_{\mathrm{S}}$ and $\mathcal{L}_{\mathrm{E}}$, respectively. Hence, the whole classification loss is formulated as: $\mathcal{L}_{\mathrm{cls}}=\mathcal{L}_{\mathrm{C}}+\mathcal{L}_{\mathrm{S}}+\mathcal{L}_{\mathrm{E}}$.

To make the action boundaries more precise, we also introduce a regression task to predict the starting and ending boundary offsets. Inspired by some object detection studies \cite{ren2015faster,law2018cornernet}, we apply 
SmoothL1 Loss \cite{girshick2015fast} ($\mathrm{SL}_1$) to our regression task, which is formulated in Eq.~\eqref{eq:loss3}:

\begin{equation}\label{eq:loss3}
  \mathcal{L}_\mathrm{reg}=\frac{1}{T_\mathrm{S}^{+}}\sum_{t\in\mathcal{U}^\mathrm{S}}\mathrm{SL}_1(o_{t}^\mathrm{S},\hat{o_{t}}^\mathrm{S}) + \frac{1}{T_\mathrm{E}^{+}}\sum_{t\in\mathcal{U}^\mathrm{E}}\mathrm{SL}_1(o_{t}^\mathrm{E},\hat{o_{t}}^\mathrm{E}),
\end{equation}
where $\mathcal{U}^\mathrm{S}$ and $\mathcal{U}^\mathrm{E}$ are the \textit{positive} regions in $\mathbf{p}^\mathrm{S}$ and $\mathbf{p}^\mathrm{E}$. $T_\mathrm{S}^{+}$ and $T_\mathrm{E}^{+}$ are the number of them. $\hat{o_{t}}^\mathrm{S}$ and $\hat{o_{t}}^\mathrm{E}$ are the predicted starting and ending offsets with their ground-truth ($o_{t}^\mathrm{S}$ and $o_{t}^\mathrm{E}$). Adding our proposed consistency constrains IntraC and InterC, the overall objective loss function is formulated in Eq.~\eqref{eq:loss4}:

\begin{equation}\label{eq:loss4}
  \mathcal{L}= \mathcal{L}_\mathrm{cls} + \mathcal{L}_\mathrm{reg} + \mathcal{L}_{\mathrm{Intra}} + \mathcal{L}_{\mathrm{Inter}}.
\end{equation}

\subsection{Inference: Proposal Generation and Classification}\label{section:postprocess}

Following the same rules in BSN~\cite{lin2018bsn} and ScaleMatters~\cite{gong2019scale}, we select the starting and ending points in terms of $\mathbf{p}^\mathrm{S}$ and $\mathbf{p}^\mathrm{E}$; then combine them to generate action proposals; finally rank these proposals and classify them with action labels. Operations are conducted sequentially:

\noindent \textbf{Proposal generation}. 
To generate action proposals, we first select the candidate starting and ending points with predicted $\mathbf{p}^\mathrm{S}$ and $\mathbf{p}^\mathrm{E}$ by two rules~\cite{lin2018bsn}: (i) start points $t$ where $p_{t}^\mathrm{S}>0.5\times(\max_{t=1}^{T}\{p_{t}^\mathrm{S}\}+\min_{t=1}^{T}\{p_{t}^\mathrm{S}\})$; (ii) start points $t$ where $p_{t-1}^{\mathrm{S}}<p_{t}^{\mathrm{S}}<p_{t+1}^{\mathrm{S}}$. The ending points are selected by the same rules. Following these two rules, we obtain starting and ending candidates which have high probability or stay at a peak position. Combining these points under a maximum action duration in training set, we obtain the candidate proposals.

\noindent \textbf{Proposal ranking}. 
To rank action proposals with a confidence score, we provide two methods: (i) directly use the product of the starting and ending probabilities, $p_{t_{s}}^\mathrm{S}\times p_{t_{e}}^\mathrm{E}$. (ii) train an additional evaluation network to score candidate proposals \cite{gong2019scale}, which is noted as $\phi(t_{s},t_{e})$. The detailed information can be found in \cite{gong2019scale}. Thus, the final confidence score for candidate proposals is $p_{t_{s}}^\mathrm{S}\times p_{t_{e}}^\mathrm{E}\times\phi(t_{s},t_{e})$.

\noindent \textbf{Redundant proposal suppression}.
After generating candidate proposals with the confidence score, we need to remove redundant proposals with high overlaps. Standard method such as soft non-maximum suppression (Soft-NMS) \cite{bodla2017soft} is used in our experiments. Soft-NMS decays the confidence score of proposals which are highly overlapped. Finally, we suppress the redundant proposals to achieve a higher recall.

\noindent \textbf{Proposal classification}. 
The last step of temporal action localization is to classify the candidate proposals. For fair comparison with other temporal localization methods, we use the same classifiers to report our action localization results. Following BSN~\cite{lin2018bsn}, we use video-level classifier in UntrimmedNet~\cite{wang2017untrimmednets} for THUMOS14 dataset. As for ActivityNet1.3 dataset, we use the video-level classification results generated by \cite{zhao2017cuhk}.

\subsection{Implementation Details}\label{section:network&loss}

\noindent \textbf{Network Design}. We build our IntraC and InterC on a succinct baseline model with all 1D Convolution layers and the detailed network architecture is shown in Table~\ref{table:model}. The input of BaseNet is extracted feature sequence $\{\mathbf{f}_{t}\}_{t=1}^{T}$ of untrimmed videos. Since untrimmed videos have various video length, we truncate or pad zeros to obtain a fixed length features of window $l_\mathrm{w}$. Through BaseNet, the output features are shared by three 2-layer ProbNets to predict probability phases ($\mathbf{p}^\mathrm{C}$, $\mathbf{p}^\mathrm{S}$, and $\mathbf{p}^\mathrm{E}$) and two RegrNets to predict starting and ending boundary offsets ($\hat{\mathbf{o}}_\mathrm{S}$ and $\hat{\mathbf{o}}_\mathrm{E}$).

\begin{wraptable}{r}{0.55\textwidth}
\small
\begin{center}
\caption{The detailed network architecture. The output of BaseNet is shared by ProbNet and RegrNet. Three ProbNets ($\times$ 3) are used to predict continuing, starting, and ending probability phases. Two RegrNets ($\times$ 2) are used to predict starting and ending offsets.}
\label{table:model}
\begin{tabular}{ccccc}
\toprule
 Name                                                                               & Layer   & Kernel & Channels & Activation \\ \hline
\multirow{2}{*}{BaseNet}                                                            & Conv1D  & 9           & 512      & ReLU       \\
                                                                                    & Conv1D  & 9           & 512      & ReLU       \\ \hline
\multirow{2}{*}{\begin{tabular}[c]{@{}c@{}}ProbNet\\ ($\times$ 3)\end{tabular}}     & Conv1D  & 5           & 256      & ReLU       \\
                                                                                    & Conv1D  & 5           & 1        & Sigmoid    \\ \hline
\multirow{2}{*}{\begin{tabular}[c]{@{}c@{}}RegrNet\\ ($\times$ 2)\end{tabular}}     & Conv1D  & 5           & 256      & ReLU       \\
                                                                                    & Conv1D  & 5           & 1        & Identity   \\ \bottomrule
\end{tabular}
\end{center}
\end{wraptable}

\noindent \textbf{Network training}. Our BaseNet, ProbNet, and RegrNet are jointly trained from scratch by multiple losses which are the classification loss ($\mathcal{L}_\mathrm{cls}$), regression loss ($\mathcal{L}_\mathrm{reg}$) and consistency losses ($\mathcal{L}_{\mathrm{Intra}}$ and $\mathcal{L}_{\mathrm{Inter}}$). We find setting the ratio of each loss component equal get relatively proper numerical values and the loss curve can converge well. As mentioned previous, to contain most action instances in a fixed observed window, the input feature length of window $l_\mathrm{w}$ is set to be $750$ for THUMOS14 and scaled to be $100$ for ActivityNet1.3.
The training process lasts for $20$ epochs with a learning rate of $10^{-3}$ in former $10$ epochs and $10^{-4}$ in latter $10$ epochs. The batch size is set to be $3$ for THUMOS14 and $16$ for the ActivityNet1.3. We use a SGD optimization method with a momentum of $0.9$ to train both datasets. In Section \ref{section:postprocess}, the additional evaluation network for proposal ranking follows the same settings in \cite{gong2019scale}.

\section{Experiments}

\subsection{Datasets and Evaluation Metrics}

\noindent \textbf{Datasets and features}.
We validate our proposed IntraC and InterC on two standard datasets: \textbf{THUMOS14} includes $413$ untrimmed videos with $20$ action classes. According to the public split, $200$ of them are used for training, and $213$ are used for testing. There are more than 15 action annotations in each video; \textbf{ActivityNet1.3} is a more considerable action localization dataset with $200$ classes annotated. The entire $19,994$ untrimmed videos are divided into training, validation, and testing sets by ratio $2$:$1$:$1$. Each video has around $1.5$ action instances.  To make a fair comparison with the previous work, we use the same two-stream features of these datasets. The two-stream features, which are provided by \cite{liu2019completeness}, are extracted by I3D network \cite{carreira2017quo} pre-trained on Kinetics.

\noindent \textbf{Metric for temporal action proposals}.
To evaluate the quality of action proposals, we use conventional metrics Average Recall (AR) with different Average Number (AN) of proposals AR@AN for action proposals. On THUMOS14 dataset, the AR is calculated under multiple IoU threshold set from $0.5$ to $1.0$ with a stride of $0.05$. 
As for ActivityNet1.3 dataset the multiple IoU threshold are from $0.5$ to $0.95$ with a stride of $0.05$. Besides, we also use the area under the AR-AN curve (AUC) to evaluate the performance. 

\noindent \textbf{Metric for temporal action localization}.
To evaluate the performance of action localization, we use mean Average Precision (mAP) metric. On THUMOS14 dataset, we report the mAP with multiple IoUs in set $\{0.3,0.4,0.5,0.6,0.7\}$. As for ActivityNet1.3 dataset, the IoU set is $\{0.5,0.7,0.95\}$. Moreover, we also report the averaged mAP where the IoU is from $0.5$ to $0.95$ with a stride of $0.05$.

\subsection{Comparison to the State-of-the-arts}

\noindent \textbf{Temporal action proposals}. We compare the temporal action proposals generated by our IntraC and InterC equipped model on THUMOS14 and ActivityNet1.3 dataset. As illustrated in Table~\ref{table:AR@AN_THUMOS14}, comparing with previous works, we can achieve the best performance especially on AR@50 metric. Our consistency losses help to generate more precise candidate starting and ending points, so we can achieve a high recall with fewer proposals. In Table~\ref{table:AR@AN_ANET}, we also achieve comparable results on ActivityNet1.3, since it is a well studied dataset.

\noindent \textbf{Temporal action localization}. Classifying the proposed proposals, we obtain the final localization results. As illustrated in Table~\ref{table:mAP_THUMOS14} and Table~\ref{table:mAP_ANET}, our method outperforms the previous studies. Especially at \textbf{high IoU settings}, we achieve significant improvements since our consistency loss can make the boundaries more precise. On THUMOS14 dataset, the mAP at IoU of $0.6$ is improved from $31.5\%$ to $38.0\%$ and the mAP at IoU of $0.7$ is improved from $21.7\%$ to $28.5\%$. On ActivityNet1.3 dataset, we can achieve the mAP to $9.21\%$ at IoU of $0.95$.

\begin{table}[t] 
\begin{minipage}{0.48\linewidth} 
\small
\centering 
\caption{Comparisons in terms of AR@AN (\%) on THUMOS14.} 
    \begin{tabular}{C{2cm}C{1cm}C{1cm}C{1cm}}
    \toprule
    Method                                  & @50   & @100   & @200   \\ \hline
    TAG~\cite{xiong2017pursuit}             & 18.55 & 29.00  & 39.41  \\
    CTAP~\cite{gao2018ctap}                 & 32.49 & 42.61  & 51.97  \\ 
    BSN~\cite{lin2018bsn}                   & 37.46 & 46.06  & 53.21  \\
    BMN~\cite{lin2019bmn}                   & 39.36 & 47.72  & 54.70  \\
    MGG~\cite{liu2019multi}                 & 39.93 & 47.75  & 54.65  \\
    TSA-Net~\cite{gong2019scale}            & 42.83 & 49.61  & 54.52  \\ \cmidrule(lr){1-1} \cmidrule(lr){2-2} \cmidrule(lr){3-3} \cmidrule(lr){4-4} 
    Ours                                    & \textbf{44.23} & \textbf{50.67}  & \textbf{55.74}  \\ \bottomrule
    \end{tabular}
\label{table:AR@AN_THUMOS14}
\end{minipage}
\begin{minipage}{0.48\linewidth} 
\small
\centering 
\caption{Comparisons in terms of mAP (\%) on THUMOS14.}
\begin{tabular}{cccccc}
    \toprule
    Method                               & 0.3   & 0.4   & 0.5   & 0.6   & 0.7   \\ \hline
    SST~\cite{buch2017sst}               & 41.2  & 31.5  & 20.0  & 0.9   & 4.7   \\
    TURN~\cite{gao2017turn}              & 46.3  & 35.3  & 24.5  & 14.1  & 6.3   \\
    BSN~\cite{lin2018bsn}                & 53.5  & 45.0  & 36.9  & 28.4  & 20.0  \\
    MGG~\cite{liu2019multi}              & 53.9  & 46.8  & 37.4  & 29.5  & 21.3  \\
    BMN~\cite{lin2019bmn}                & \textbf{56.0}  & 47.4  & 38.8  & 29.7  & 20.5  \\
    TSA-Net~\cite{gong2019scale}         & 53.2  & 48.1  & 41.5  & 31.5  & 21.7  \\ \cmidrule(lr){1-1} \cmidrule(lr){2-2} \cmidrule(lr){3-3} \cmidrule(lr){4-4} \cmidrule(lr){5-5} \cmidrule(lr){6-6}
    Ours                                 & 53.9  & \textbf{50.7}  & \textbf{45.4}  & \textbf{38.0}  & \textbf{28.5}  \\ \bottomrule
    \end{tabular}
\label{table:mAP_THUMOS14}
\end{minipage} 
\end{table}

\begin{table}[t] 
\begin{minipage}{0.48\linewidth} 
\small
\centering 
\caption{Comparisons in terms of AUC and AR@100 (\%) on ActivityNet1.3.} 
    \begin{tabular}{C{1.8cm}C{1.6cm}C{1.6cm}}      
    \toprule
    Method                          & AUC               &AR@100     \\ \hline
    TCN \cite{dai2017temporal}      & 59.58             & -         \\
    CTAP \cite{gao2018ctap}         & 65.72             & 73.17     \\
    BSN \cite{lin2018bsn}           & 66.17             & 74.16     \\
    MGG \cite{liu2019multi}         & 66.43             & 74.54     \\
    \cmidrule(lr){1-1} \cmidrule(lr){2-2} \cmidrule(lr){3-3}
    Ours                            & \textbf{66.51}    & \textbf{75.27} \\
    \bottomrule
    \end{tabular}
\label{table:AR@AN_ANET}
\end{minipage}
\begin{minipage}{0.48\linewidth} 
\small
\centering 
\caption{Comparisons in terms of mAP (\%) on ActivityNet1.3 (val). ``Average'' is caculated at the IoU of $\{0.5:0.05:0.95\}$.}
\begin{tabular}{C{1.4cm}C{0.9cm}C{0.9cm}C{0.9cm}C{1.2cm}}
        \toprule
        Method                           & 0.5   & 0.7   & 0.95  & Average     \\ \hline
        CDC \cite{shou2017cdc}           & 43.83 & 25.88 & 0.21  & 22.77                   \\
        SSN \cite{zhao2017temporal}      & 39.12 & 23.48 & 5.49  & 23.98                   \\
        BSN \cite{lin2018bsn}            & \textbf{46.45} & 29.96 & 8.02  & 30.03                   \\ \cmidrule(lr){1-1} \cmidrule(lr){2-2} \cmidrule(lr){3-3} \cmidrule(lr){4-4} \cmidrule(lr){5-5}
        Ours                             & 43.47 & \textbf{33.91} & \textbf{9.21}  & \textbf{30.12}                   \\ \bottomrule
        \end{tabular}
\label{table:mAP_ANET}
\end{minipage} 
\end{table}

\noindent \textbf{Generalizing IntraC\&InterC to Other Algorithms.} Our proposed two consistency losses, \emph{i.e.}, IntraC and InterC, are effective in generating the probability phases of continuing, starting, and ending. To prove these consistency losses are valid for other network architecture and framework in TAL, we introduce them to TSA-Net~\cite{gong2019scale} and PGCN~\cite{zeng2019graph}, respectively. \textbf{TSA-Net}~\cite{gong2019scale} designed a multi-scale architecture to predict probability phases of continuing, starting and ending. We introduce our IntraC and InterC to their multi-scale networks, TSA-Net-small, TSA-Net-medium, and TSA-Net-large, respectively. As illustrated in Table~\ref{table:AR_ScaleMatters}, our IntraC and InterC significantly outperforms the baseline models on all three network architectures. \textbf{PGCN}~\cite{zeng2019graph} explore the proposal-proposal relations using Graph Convolutional Networks \cite{kipf2017semi} (GCN) to localize action instances. This framework builds upon the prepared proposals from BSN \cite{lin2018bsn} method. We introduce our two consistency losses to generated candidate proposals for PGCN framework. As illustrated in Table~\ref{table:mAP_PGCN}, introducing IntraC and InterC to PGCN also improves the localization performance.

\begin{table}[t]
\small
\begin{center}
\caption{Generalizing IntraC\&InterC to multi-scale TSA-Net~\cite{gong2019scale} in terms of AR@AN (\%) on THUMOS14. $\ast$ indicates the results that are implemented by ours.}
\label{table:AR_ScaleMatters}
\begin{tabular}{C{4.2cm}C{2.2cm}C{2.2cm}C{2.2cm}}
\toprule
TSA-Net                                          & AR@50                  & AR@100              & AR@200          \\ \hline
Small (Small\textsuperscript{*})                 & 37.72 (38.32)          & 45.85 (46.15)       & 52.03 (52.39)   \\
Small\textsuperscript{*} + IntraC\&InterC        & \textbf{39.73}         & \textbf{47.69}      & \textbf{53.48}  \\ \hline
Medium (Medium\textsuperscript{*})               & 37.77 (39.20)          & 45.01 (47.17)       & 50.38 (53.46)   \\
Medium\textsuperscript{*} + IntraC\&InterC       & \textbf{40.05}         & \textbf{47.53}      & \textbf{53.88}  \\ \hline
Large (Large\textsuperscript{*})                 & 36.07 (37.91)          & 44.28 (45.89)       & 50.80 (52.36)   \\
Large\textsuperscript{*} + IntraC\&InterC        & \textbf{39.68}         & \textbf{47.47}      & \textbf{53.50}  \\\bottomrule
\end{tabular}
\end{center}
\end{table}

\begin{table}[t]
\small
\begin{center}
\caption{Generalizing IntraC\&InterC to PGCN~\cite{zeng2019graph} in terms of mAP (\%) on THUMOS14. $\ast$ indicates the results that are implemented by ours.}
\label{table:mAP_PGCN}
\begin{tabular}{C{4.2cm}C{1.2cm}C{1.2cm}C{1.2cm}C{1.2cm}C{1.2cm}}
\toprule
Method                                       & 0.1             & 0.2             & 0.3             & 0.4             & 0.5             \\ \hline
PGCN                                         & 69.50           & 67.80           & 63.60           & 57.80           & 49.10  \\
PGCN\textsuperscript{*}                      & 69.26           & 67.76           & 63.73           & 58.82           & 48.88           \\
PGCN\textsuperscript{*} + IntraC\&InterC     & \textbf{71.83}  & \textbf{70.31}  & \textbf{66.29}  & \textbf{60.99}  & \textbf{50.10}           \\
\bottomrule
\end{tabular}
\end{center}
\end{table}

\subsection{Ablation Studies}

As mentioned in dataset description, THUMOS has 10 times action instances per video than ActivityNet (only has 1.5 action instances per video) and THUMOS video also contains a larger portion of background. More instances and more background are challenge for detection task. Thus we conduct following detailed ablation studies on THUMOS14 dataset to explore how these constrains, IntraC and InterC, improve the quality of temporal action proposals.

\noindent \textbf{Effectiveness of IntraC.} As illustrated in Table~\ref{table:ICEC} ``Intra Consistency'', we compare the components of IntraC in terms of the AR@AN. The IntraC is introduced to continuing probability phase ($\mathcal{L}_{\mathrm{Intra}^\mathrm{C}}$), starting probability phase ($\mathcal{L}_{\mathrm{Intra}^\mathrm{S}}$), and ending probability phase ($\mathcal{L}_{\mathrm{Intra}^\mathrm{E}}$). Compared with the baseline result without any consistency losses, introducing continuing $\mathcal{L}_{\mathrm{Intra}^\mathrm{C}}$ or starting $\mathcal{L}_{\mathrm{Intra}^\mathrm{S}}$ and ending $\mathcal{L}_{\mathrm{Intra}^\mathrm{E}}$ can both achieve better results. Combined all three IntraC losses, the AR@50 is improved from $39.02\%$ to $41.91\%$.

\noindent \textbf{Effectiveness of InterC.} As illustrated in Table~\ref{table:ICEC} ``Inter Consistency'', we compare the components of InterC losses in terms of the AR@AN. The InterC is introduced between continue-start (C\&S) and continue-end (C\&E). InterC on C\&S (C\&E) makes the consistency between the starting phase (ending phase) and the derivative of continuing phase, which can suppress the false positives only observed from a single probability phase. Only introducing InterC to C\&S or C\&E obtains around $1\%$ absolute improvement on AR@50. When combined C\&S and C\&E, it can improve $2.21\%$ on AR@50.

\noindent \textbf{Combining IntraC\&InterC.} As illustrated in Table~\ref{table:ICEC} ``All Consistency'', we compare the IntraC and InterC losses in terms of the AR@AN. Both the IntraC and InterC independently achieve more than $2\%$ absolute improvement on AR@50. When combined IntraC and InterC, the AR@50 is improved from $39.02\%$ to $42.63\%$. Consistency inside each probability phase and between them are coupled, which leads to a positive feedback. It means when we get the better probability phase that fits the IntraC settings, the potential constraint of InterC is more appropriate between three probability phases, and vise versa.

\begin{table}[t]
\small
\begin{center}
\caption{Ablation studies on Intra-phase Consistency and Inter-phase Consistency in terms of AR@AN (\%) on THUMOS14. The baseline model is define in Table~\ref{table:model}. }
\label{table:ICEC}
\renewcommand{\arraystretch}{1.0}
 \setlength{\tabcolsep}{1.5mm}{ 
  \scalebox{0.9}{
\begin{tabular}{C{0.8cm}C{0.8cm}C{0.8cm}C{0.8cm}C{0.8cm}C{0.8cm}C{1.8cm}C{1.8cm}C{1.8cm}}
\toprule
              &              &             &             &            &                                      & AR@50       & AR@100       & AR@200      \\ \hline
\multicolumn{6}{c}{Baseline}                                                                                 & 39.02       & 46.26        & 53.09       \\ \hline \hline
\multicolumn{2}{c}{Continue} & \multicolumn{2}{c}{Start} & \multicolumn{2}{c}{End}                           & \multicolumn{3}{c}{Intra Consistency} \\ \hline
\multicolumn{2}{c}{\checkmark}        & \multicolumn{2}{c}{}      & \multicolumn{2}{c}{}                     & 40.46       & 47.85        & 53.87        \\
\multicolumn{2}{c}{}         & \multicolumn{2}{c}{\checkmark}     & \multicolumn{2}{c}{\checkmark}           & 40.86       & 48.26        & 54.16        \\
\multicolumn{2}{c}{\checkmark}        & \multicolumn{2}{c}{\checkmark}     & \multicolumn{2}{c}{\checkmark}  & 41.91       & 49.06        & 54.82       \\ \hline \hline
\multicolumn{3}{c}{C\&S}                   & \multicolumn{3}{c}{C\&E}                                        & \multicolumn{3}{c}{Inter Consistency} \\ \hline
\multicolumn{3}{c}{\checkmark}                      & \multicolumn{3}{c}{}                                   & 40.21       & 47.30        & 53.38       \\
\multicolumn{3}{c}{}                       & \multicolumn{3}{c}{\checkmark}                                  & 40.64       & 47.85        & 54.01       \\
\multicolumn{3}{c}{\checkmark}                      & \multicolumn{3}{c}{\checkmark}                         & 41.23       & 48.81        & 54.47       \\ \hline \hline
\multicolumn{3}{c}{IntraC}                     & \multicolumn{3}{c}{InterC}                                          & \multicolumn{3}{c}{Intra\&Inter Consistency}      \\ \hline
\multicolumn{3}{c}{\checkmark}                      & \multicolumn{3}{c}{}                                   & 41.91       & 49.06        & 54.82       \\
\multicolumn{3}{c}{}                       & \multicolumn{3}{c}{\checkmark}                                  & 41.23       & 48.81        & 54.47       \\
\multicolumn{3}{c}{\checkmark}                      & \multicolumn{3}{c}{\checkmark}                         & \textbf{42.63}    & \textbf{49.85}       & \textbf{55.32}       \\ \bottomrule
\end{tabular}}}
\end{center}
\end{table}

\noindent \textbf{Effectiveness of kernel size and layers.} The scale of the receptive field is crucial in temporal action localization tasks. So we explore different scales of receptive field by adjusting the number of layers and the kernel size of the BaseNet. As illustrated in Table~\ref{table:structure}, we compare results between different kernel sizes and layers in terms of the AR@AN. Deeper layers and larger kernel sizes often lead to a better performance, but using too many layers and/or an over-large kernel size often incurs over-fitting. We also conduct the experiments using different layers with a kernel size of 9 and find that a 2-layer network performs best, so we use this option in the main experiments. This implies that probably increasing the depth is not the best choice here.
%though we still hope the receptive field to be sufficiently large.

\begin{table}[t]
\small
\begin{center}
\caption{Ablation studies on model structures in terms of AR@AN (\%) on THUMOS14. All numbers are the averaged value in the last $10$ epochs.}
\label{table:structure}
\renewcommand{\arraystretch}{1.0}
 \setlength{\tabcolsep}{1.5mm}{ 
  \scalebox{0.9}{
\begin{tabular}{C{2.4cm}C{2.4cm}C{1.8cm}C{1.8cm}C{1.8cm}}
\toprule
Layers      & Kernel Size   & AR@50             & AR@100               & AR@200 \\ \hline
2           & 5             & 40.98             & 48.51                & 54.64  \\
3           & 5             & 41.56             & 49.02                & 54.88  \\
4           & 5             & 41.68             & 48.93                & 54.91  \\ 
5           & 5             & 40.98             & 48.14                & 54.29  \\ \hline
2           & 3             & 39.54             & 47.61                & 53.84  \\
2           & 5             & 40.98             & 48.51                & 54.64  \\
2           & 7             & 41.49             & 49.16                & 55.17  \\
2           & 9             & \textbf{42.63}    & \textbf{49.85}       & \textbf{55.32}  \\
2           & 11            & 42.48             & 49.32                & 54.97  \\
2           & 13            & 42.17             & 49.41                & 55.21  \\ \bottomrule
% 2           & 15            & 41.69             & 48.84                & 54.26  \\ \bottomrule
\end{tabular}}}
\end{center}
\end{table}

\noindent \textbf{Effectiveness of proposal scoring.} As mentioned in Section \ref{section:postprocess}, we compare two methods for scoring proposals. Once we get proposals of an untrimmed video, a proper ranking method with convincing scores can achieve the high recall with fewer proposals. As illustrated in Table~\ref{table:scoring}, we compare two scoring functions, $p_{t_{s}}^\mathrm{S}\times p_{t_{e}}^\mathrm{E}$ and $p_{t_{s}}^\mathrm{S}\times p_{t_{e}}^\mathrm{E}\times\phi(t_{s},t_{e})$. Directly using starting and ending probability at boundaries is simple and effective, however, training a new evaluation network \cite{lin2018bsn,gong2019scale} to evaluate the confidence of proposals can further improve the performance by a significant margin.

\begin{table}[t]
\small
\begin{center}
\caption{Ablation studies on proposal scoring in terms of AR@AN (\%) on THUMOS14. Experiments are based on 2 ``Layers'' and 9 ``Kernel Size'' model in Table~\ref{table:structure}. All numbers are the averaged value in the last $10$ epochs. }
\label{table:scoring}
\renewcommand{\arraystretch}{1.0}
 \setlength{\tabcolsep}{1.5mm}{ 
  \scalebox{0.9}{
\begin{tabular}{C{4cm}C{1.8cm}C{1.8cm}C{1.8cm}}
\toprule
Proposal Scoring                                                           & AR@50     & AR@100    & AR@200 \\ \hline
$p_{t_{s}}^\mathrm{S}\times p_{t_{e}}^\mathrm{E}$                          & 42.63     & 49.85     & 55.32  \\
$p_{t_{s}}^\mathrm{S}\times p_{t_{e}}^\mathrm{E}\times\phi(t_{s},t_{e})$   & \textbf{44.23}     & \textbf{50.67}     & \textbf{55.74}  \\
\bottomrule
\end{tabular}}}
\end{center}
\end{table}

\begin{figure*}[t]
\begin{center}
  \includegraphics[width=0.49\linewidth]{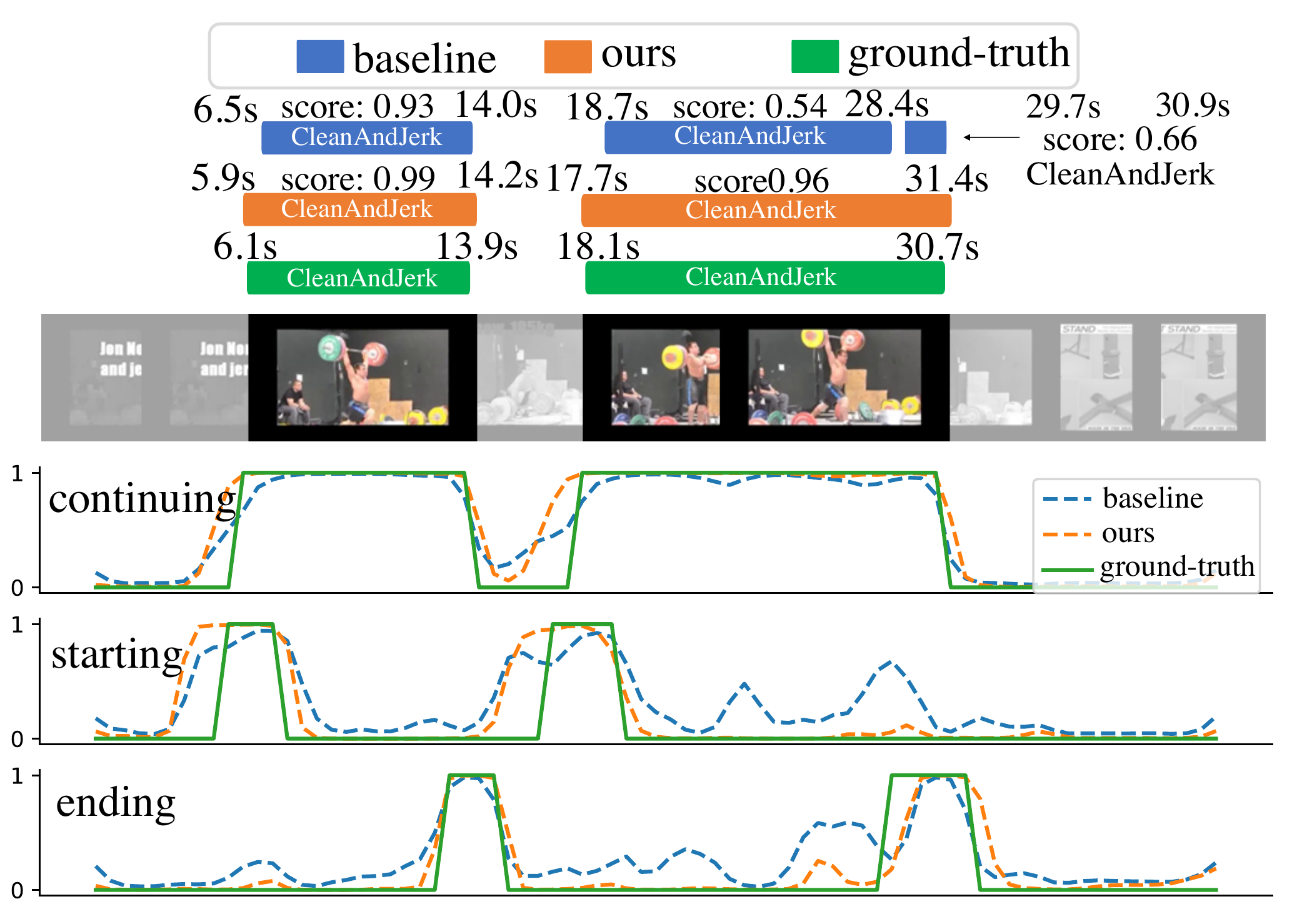}
  \includegraphics[width=0.49\linewidth]{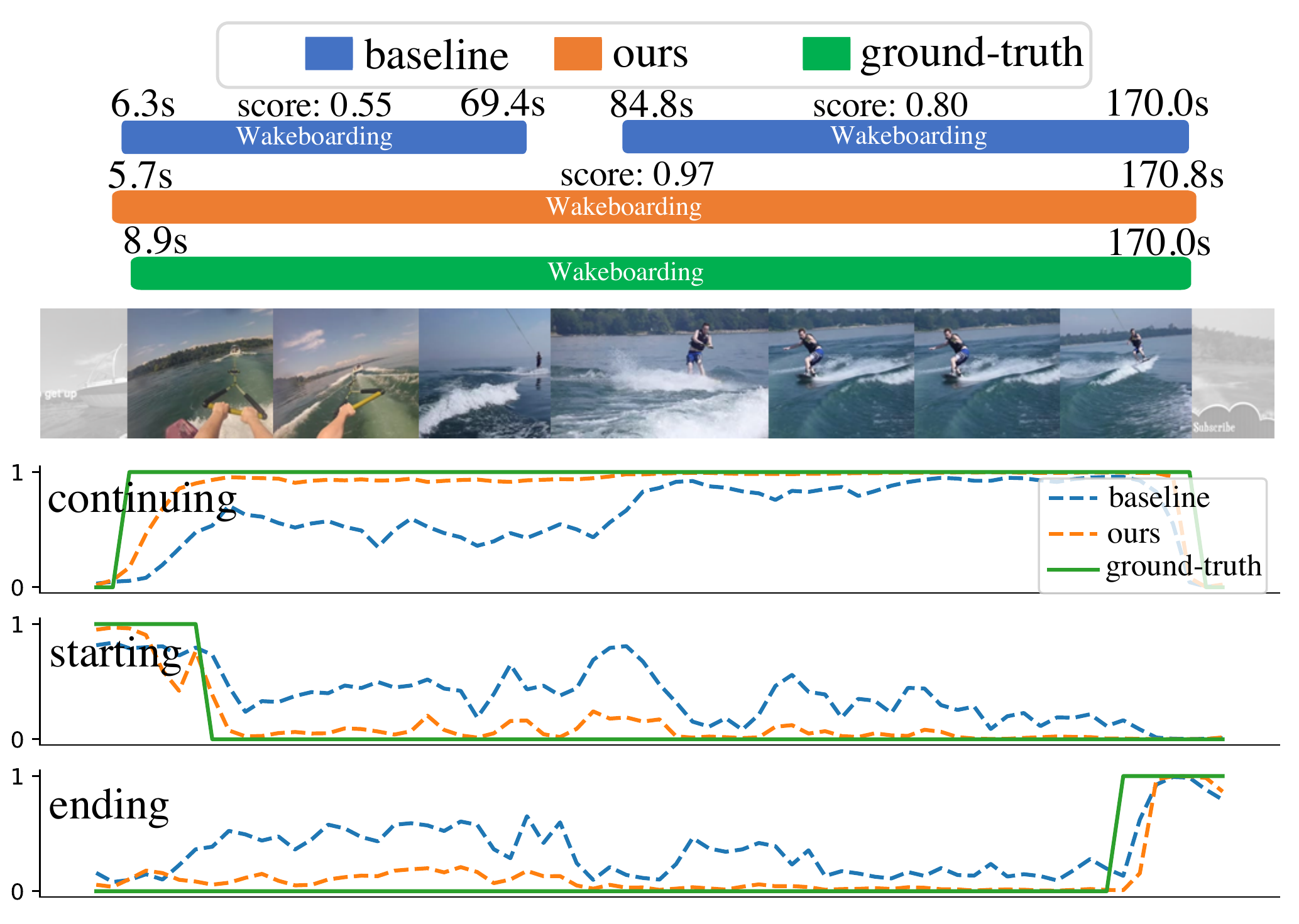}
\end{center}
\caption{Qualitative results on THUMOS14 (left) and ActivityNet1.3 (right) datasets. ``green'' lines are ground-truth, ``blue'' lines are predicted phases by baseline model and ``orange'' lines are optimized with IntraC and InterC regularization terms. }
\label{fig:visulization}
\end{figure*}

\subsection{Visualization}

As illustrated in Figure~\ref{fig:visulization}, we visualize some examples on both datasets. Comparing the predicted $\mathbf{p}^\mathrm{C}$, $\mathbf{p}^\mathrm{S}$, and $\mathbf{p}^\mathrm{E}$ with or without the IntraC and InterC regularization, we find our proposed IntraC and InterC indeed make each predicted phase becomes stable inside \textit{foreground} and \textit{background} regions. Besides, some false positives in $\mathbf{p}^\mathrm{S}$ and $\mathbf{p}^\mathrm{E}$ are suppressed by their context information, so that we can remove many candidate proposals of poor quality via these wrong starting and ending points. \emph{e.g.}, the second action ``CleanAndJerk'' and the action ``Wakeboarding'' are separate by false positive starting point in baseline model. The visualization results show that only introducing binary classification labels is hard to optimize these probability phases, since it discards the potential constraints between the different temporal locations and action phases. 
We also perform regularization using the smoothness assumption, \textit{i.e.}, using a Gaussian kernel to penalize local inconsistenies within $\mathbf{p}^\mathrm{C}$, $\mathbf{p}^\mathrm{S}$, and $\mathbf{p}^\mathrm{E}$. In experiments, this kinds of regularization does not necessarily push the positive scores to 1 and negative scores to 0, and we believe smoothness might be useful in the unsupervised or weakly-supervised TAL scenarios.

\subsection{Discussion: the Upper Bounds of TAL}

Most temporal action localization method can be divided into the following procedures, (i) generating proposals, (ii) ranking proposals, and (iii) classifying proposals. Which one is most awaiting to improve for the intending researchful keystone? We introduce two types of oracle information to reveal the performance gap between the different upper bounds. As illustrated in Table~\ref{table:oracle}, $O_{\mathrm{rank}}$ means that each candidate proposal is ranked by the max IoU score with all ground-truth action instances. $O_{\mathrm{cls}}$ means that the ground-truth action labels are assigned to candidate proposals. When introducing $O_{\mathrm{rank}}$ or/and $O_{\mathrm{cls}}$ to our action localization baseline, it is worth to notice that proposal classification has been well solved since there is a small gap when introducing $O_{\mathrm{cls}}$. However, when introducing the oracle ranking information $O_{\mathrm{rank}}$, the upper bound can improve a lot from $53.9\%$ to $66.4\%$ in terms of mAP at IoU of $0.3$. That means there is a significant untapped opportunity in how to rank the action proposals.

\begin{table}[t]
\small
\begin{center}
\caption{Introducing oracle information to TAL in terms of mAP (\%) on THUMOS14. $O_{\mathrm{rank}}$ is ground-truth rank information and $O_{\mathrm{cls}}$ uses ground-truth class label.}
\label{table:oracle}
\begin{tabular}{C{1.2cm}C{1.2cm}C{1.2cm}C{1.2cm}C{1.2cm}C{1.2cm}C{1.2cm}}
\toprule
$O_{\mathrm{rank}}$           & $O_{\mathrm{cls}}$         & 0.3                  & 0.4                  & 0.5                  & 0.6                  & 0.7              \\ \hline
                     &                   & 53.9                 & 50.7                 & 45.4                 & 38.0                 & 28.5             \\
                     & \checkmark        & 57.1                 & 53.2                 & 47.3                 & 39.3                 & 29.5             \\
 \checkmark          &                   & 66.4                 & 65.4                 & 63.8                 & 59.9                 & 52.7             \\
 \checkmark          & \checkmark        & \textbf{72.1}        & \textbf{70.9}        & \textbf{68.8}        & \textbf{64.1}        & \textbf{55.6}    \\ \bottomrule
\end{tabular}
\end{center}
\end{table}

\section{Conclusions}
In this paper, we investigate the problem that frame-level probability phases of starting, continuing, and ending are not self-consistent in the bottom-up TAL approach. 
Our research reveals that state-of-the-art video analysis algorithms, though supervised with classification labels, mostly have a limited understanding in the temporal dimension, which can lead to undesired properties, \textit{e.g.}, inconsistency or discontinuity. 
To alleviate this problem, we propose two consistency losses (IntraC and InterC) which can mutually regularize the learning process. Experiments reveal that our approach improves the performance of temporal action localization both quantitatively and qualitatively.

Our work reveals that introducing priors for self-regularization is important for learning from high-dimensional data (\textit{e.g.}, videos). We will continue along this direction in the future, and explore the possibility of learning such priors from self-supervised data, \textit{e.g.}, unlabeled videos.

% \section*{Acknowledgements}
\noindent \textbf{Acknowledgements}
This work is supported by the National Key Research and Development Program of China (No. 2019YFB1804304), SHEITC (No. 2018-RGZN-02046), 111 plan (No. BP0719010),  and STCSM (No. 18DZ2270700), and State Key Laboratory of UHD Video and Audio Production and Presentation.

% ---- Bibliography ----
%
% BibTeX users should specify bibliography style 'splncs04'.
% References will then be sorted and formatted in the correct style.
%
\bibliographystyle{splncs04}
\bibliography{egbib}

\begin{thebibliography}{10}
\providecommand{\url}[1]{\texttt{#1}}
\providecommand{\urlprefix}{URL }
\providecommand{\doi}[1]{https://doi.org/#1}

\bibitem{abu2016youtube}
Abu-El-Haija, S., Kothari, N., Lee, J., Natsev, P., Toderici, G., Varadarajan,
  B., Vijayanarasimhan, S.: Youtube-8m: A large-scale video classification
  benchmark. In: arXiv preprint arXiv:1609.08675 (2016)

\bibitem{bodla2017soft}
Bodla, N., Singh, B., Chellappa, R., Davis, L.S.: Soft-nms--improving object
  detection with one line of code. In: Proceedings of the International
  Conference on Computer Vision (ICCV). pp. 5561--5569 (2017)

\bibitem{buch2017sst}
Buch, S., Escorcia, V., Shen, C., Ghanem, B., Carlos~Niebles, J.: Sst:
  Single-stream temporal action proposals. In: Proceedings of the Conference on
  Computer Vision and Pattern Recognition (CVPR). pp. 2911--2920 (2017)

\bibitem{caba2016fast}
Caba~Heilbron, F., Carlos~Niebles, J., Ghanem, B.: Fast temporal activity
  proposals for efficient detection of human actions in untrimmed videos. In:
  Proceedings of the Conference on Computer Vision and Pattern Recognition
  (CVPR). pp. 1914--1923 (2016)

\bibitem{caba2015activitynet}
Caba~Heilbron, F., Escorcia, V., Ghanem, B., Carlos~Niebles, J.: Activitynet: A
  large-scale video benchmark for human activity understanding. In: Proceedings
  of the IEEE Conference on Computer Vision and Pattern Recognition (CVPR). pp.
  961--970 (2015)

\bibitem{carreira2017quo}
Carreira, J., Zisserman, A.: Quo vadis, action recognition? a new model and the
  kinetics dataset. In: Proceedings of the Conference on Computer Vision and
  Pattern Recognition (CVPR). pp. 6299--6308 (2017)

\bibitem{chao2018rethinking}
Chao, Y.W., Vijayanarasimhan, S., Seybold, B., Ross, D.A., Deng, J.,
  Sukthankar, R.: Rethinking the faster r-cnn architecture for temporal action
  localization. In: Proceedings of the IEEE Conference on Computer Vision and
  Pattern Recognition (CVPR). pp. 1130--1139 (2018)

\bibitem{dai2017temporal}
Dai, X., Singh, B., Zhang, G., Davis, L.S., Qiu~Chen, Y.: Temporal context
  network for activity localization in videos. In: Proceedings of the IEEE
  International Conference on Computer Vision (ICCV). pp. 5793--5802 (2017)

\bibitem{escorcia2016daps}
Escorcia, V., Heilbron, F.C., Niebles, J.C., Ghanem, B.: Daps: Deep action
  proposals for action understanding. In: Proceedings of the European
  Conference on Computer Vision (ECCV). pp. 768--784. Springer (2016)

\bibitem{gan2016webly}
Gan, C., Sun, C., Duan, L., Gong, B.: Webly-supervised video recognition by
  mutually voting for relevant web images and web video frames. In: European
  Conference on Computer Vision (2016)

\bibitem{gan2016you}
Gan, C., Yao, T., Yang, K., Yang, Y., Mei, T.: You lead, we exceed: Labor-free
  video concept learning by jointly exploiting web videos and images. In:
  Proceedings of the IEEE Conference on Computer Vision and Pattern Recognition
  (2016)

\bibitem{gao2018ctap}
Gao, J., Chen, K., Nevatia, R.: Ctap: Complementary temporal action proposal
  generation. In: Proceedings of the European Conference on Computer Vision
  (ECCV). pp. 68--83 (2018)

\bibitem{gao2017turn}
Gao, J., Yang, Z., Chen, K., Sun, C., Nevatia, R.: Turn tap: Temporal unit
  regression network for temporal action proposals. In: Proceedings of the
  International Conference on Computer Vision (ICCV). pp. 3628--3636 (2017)

\bibitem{girshick2015fast}
Girshick, R.: Fast r-cnn. In: Proceedings of the International Conference on
  Computer Vision (ICCV). pp. 1440--1448 (2015)

\bibitem{gong2019scale}
Gong, G., Zheng, L., Bai, K., Mu, Y.: Scale matters: Temporal scale aggregation
  network for precise action localization in untrimmed videos. In:
  International Conference on Multimedia and Expo (ICME). pp.~1--6 (2020)

\bibitem{jiang2014thumos}
Jiang, Y.G., Liu, J., Zamir, A.R., Toderici, G., Laptev, I., Shah, M.,
  Sukthankar, R.: Thumos challenge: Action recognition with a large number of
  classes (2014)

\bibitem{karpathy2014large}
Karpathy, A., Toderici, G., Shetty, S., Leung, T., Sukthankar, R., Fei-Fei, L.:
  Large-scale video classification with convolutional neural networks. In:
  Proceedings of the Conference on Computer Vision and Pattern Recognition
  (CVPR). pp. 1725--1732 (2014)

\bibitem{kay2017kinetics}
Kay, W., Carreira, J., Simonyan, K., Zhang, B., Hillier, C., Vijayanarasimhan,
  S., Viola, F., Green, T., Back, T., Natsev, P., et~al.: The kinetics human
  action video dataset. In: arXiv preprint arXiv:1705.06950 (2017)

\bibitem{kipf2017semi}
Kipf, T.N., Welling, M.: Semi-supervised classification with graph
  convolutional networks. In: International Conference on Learning
  Representations (ICLR). pp. 1--14 (2017)

\bibitem{law2018cornernet}
Law, H., Deng, J.: Cornernet: Detecting objects as paired keypoints. In:
  Proceedings of the European Conference on Computer Vision (ECCV). pp.
  734--750 (2018)

\bibitem{lin2019tsm}
Lin, J., Gan, C., Han, S.: Tsm: Temporal shift module for efficient video
  understanding. In: ICCV (2019)

\bibitem{lin2019bmn}
Lin, T., Liu, X., Li, X., Ding, E., Wen, S.: Bmn: Boundary-matching network for
  temporal action proposal generation. In: Proceedings of the International
  Conference on Computer Vision (ICCV) (2019)

\bibitem{lin2018bsn}
Lin, T., Zhao, X., Su, H., Wang, C., Yang, M.: Bsn: Boundary sensitive network
  for temporal action proposal generation. In: Proceedings of the European
  Conference on Computer Vision (ECCV). pp. 3--19 (2018)

\bibitem{liu2019completeness}
Liu, D., Jiang, T., Wang, Y.: Completeness modeling and context separation for
  weakly supervised temporal action localization. In: Proceedings of the
  Conference on Computer Vision and Pattern Recognition (CVPR). pp. 1298--1307
  (2019)

\bibitem{liu2019multi}
Liu, Y., Ma, L., Zhang, Y., Liu, W., Chang, S.F.: Multi-granularity generator
  for temporal action proposal. In: Proceedings of the Conference on Computer
  Vision and Pattern Recognition (CVPR). pp. 3604--3613 (2019)

\bibitem{monfort2019moments}
Monfort, M., Andonian, A., Zhou, B., Ramakrishnan, K., Bargal, S.A., Yan, Y.,
  Brown, L., Fan, Q., Gutfreund, D., Vondrick, C., et~al.: Moments in time
  dataset: one million videos for event understanding. In: IEEE transactions on
  Pattern Analysis and Machine Intelligence (T-PAMI). IEEE (2019)

\bibitem{zhao2020u2s}
Peisen, Z., Lingxi, X., Ya, Z., Qi, T.: Universal-to-specific framework for
  complex action recognition. In: arXiv preprint arXiv:2007.06149 (2020)

\bibitem{qiu2017learning}
Qiu, Z., Yao, T., Mei, T.: Learning spatio-temporal representation with
  pseudo-3d residual networks. In: Proceedings of the IEEE International
  Conference on Computer Vision (ICCV). pp. 5534--5542. IEEE (2017)

\bibitem{redmon2016you}
Redmon, J., Divvala, S., Girshick, R., Farhadi, A.: You only look once:
  Unified, real-time object detection. In: Proceedings of the IEEE Conference
  on Computer Vision and Pattern Recognition (CVPR). pp. 779--788 (2016)

\bibitem{ren2015faster}
Ren, S., He, K., Girshick, R., Sun, J.: Faster r-cnn: Towards real-time object
  detection with region proposal networks. In: Advances in Neural Information
  Processing Systems (NeurIPS). pp. 91--99 (2015)

\bibitem{shou2017cdc}
Shou, Z., Chan, J., Zareian, A., Miyazawa, K., Chang, S.F.: Cdc:
  Convolutional-de-convolutional networks for precise temporal action
  localization in untrimmed videos. In: Proceedings of the IEEE Conference on
  Computer Vision and Pattern Recognition (CVPR). pp. 5734--5743 (2017)

\bibitem{shou2016temporal}
Shou, Z., Wang, D., Chang, S.F.: Temporal action localization in untrimmed
  videos via multi-stage cnns. In: Proceedings of the Conference on Computer
  Vision and Pattern Recognition (CVPR). pp. 1049--1058 (2016)

\bibitem{tran2015learning}
Tran, D., Bourdev, L., Fergus, R., Torresani, L., Paluri, M.: Learning
  spatiotemporal features with 3d convolutional networks. In: Proceedings of
  the IEEE International Conference on Computer Vision (ICCV). pp. 4489--4497.
  IEEE (2015)

\bibitem{tran2018closer}
Tran, D., Wang, H., Torresani, L., Ray, J., LeCun, Y., Paluri, M.: A closer
  look at spatiotemporal convolutions for action recognition. In: Proceedings
  of the IEEE Conference on Computer Vision and Pattern Recognition (CVPR). pp.
  6450--6459 (2018)

\bibitem{wang2018appearance}
Wang, L., Li, W., Li, W., Van~Gool, L.: Appearance-and-relation networks for
  video classification. In: Proceedings of the IEEE Conference on Computer
  Vision and Pattern Recognition (CVPR). pp. 1430--1439 (2018)

\bibitem{wang2017untrimmednets}
Wang, L., Xiong, Y., Lin, D., Van~Gool, L.: Untrimmednets for weakly supervised
  action recognition and detection. In: Proceedings of the Conference on
  Computer Vision and Pattern Recognition (CVPR). pp. 4325--4334 (2017)

\bibitem{xie2018rethinking}
Xie, S., Sun, C., Huang, J., Tu, Z., Murphy, K.: Rethinking spatiotemporal
  feature learning: Speed-accuracy trade-offs in video classification. In:
  Proceedings of the European Conference on Computer Vision (ECCV). pp.
  305--321 (2018)

\bibitem{xiong2017pursuit}
Xiong, Y., Zhao, Y., Wang, L., Lin, D., Tang, X.: A pursuit of temporal
  accuracy in general activity detection. In: arXiv preprint arXiv:1703.02716
  (2017)

\bibitem{xu2017rc3d}
Xu, H., Das, A., Saenko, K.: R-c3d: Region convolutional 3d network for
  temporal activity detection. In: Proceedings of the IEEE International
  Conference on Computer Vision (ICCV). pp. 5783--5792 (2017)

\bibitem{yuan2017temporal}
Yuan, Z., Stroud, J.C., Lu, T., Deng, J.: Temporal action localization by
  structured maximal sums. In: Proceedings of the IEEE Conference on Computer
  Vision and Pattern Recognition (CVPR). pp. 3684--3692 (2017)

\bibitem{zeng2019graph}
Zeng, R., Huang, W., Tan, M., Rong, Y., Zhao, P., Huang, J., Gan, C.: Graph
  convolutional networks for temporal action localization. In: Proceedings of
  the International Conference on Computer Vision (ICCV) (2019)

\bibitem{zhao2017cuhk}
Zhao, Y., Zhang, B., Wu, Z., Yang, S., Zhou, L., Yan, S., Wang, L., Xiong, Y.,
  Lin, D., Qiao, Y., et~al.: Cuhk \& ethz \& siat submission to activitynet
  challenge 2017. In: arXiv preprint arXiv:1710.08011 (2017)

\bibitem{zhao2017temporal}
Zhao, Y., Xiong, Y., Wang, L., Wu, Z., Tang, X., Lin, D.: Temporal action
  detection with structured segment networks. In: Proceedings of the IEEE
  International Conference on Computer Vision (ICCV). pp. 2914--2923 (2017)

\end{thebibliography}

\end{document}